\documentclass[runningheads]{llncs}
\usepackage{wrapfig}
\usepackage[T1]{fontenc}
\usepackage{graphicx}
\usepackage{amsmath}
\usepackage{bbm}
\usepackage{bm}
\usepackage{amsmath,epsfig}
\usepackage{balance}
\usepackage{booktabs}
\usepackage{graphicx}
\usepackage{cite}
\usepackage{subfigure}
\usepackage{multirow}
\usepackage{hyperref}
\usepackage{amsmath}
\usepackage{float}
\usepackage{caption}
\usepackage{graphicx}
\usepackage{tabularx}
\usepackage{adjustbox}
\usepackage{floatflt}
\usepackage{caption}
\usepackage{tabularx}
\usepackage[misc]{ifsym}

\begin{document}

\title{Can LLMs' Tuning Methods Work in Medical Multimodal Domain?}

\titlerunning{Can LLMs' Tuning Methods Work in Medical Multimodal Domain?}
\authorrunning{J. Chen et al.}

\author{Jiawei Chen\inst{1,2 \ast}
\quad Yue Jiang\inst{1,2 \ast}
\quad Dingkang Yang\inst{1,2}
\quad Mingcheng Li\inst{1,2}
\\
Jinjie Wei\inst{1,2}
\quad Ziyun Qian\inst{1,2}
\quad Lihua Zhang\inst{1,2,3,4}$^{\textrm{\Letter}}$
}

\institute{Academy for Engineering and Technology, Fudan University \and
Cognition and Intelligent Technology Laboratory (CIT Lab), Institute of Meta-Medical, Fudan University \and
Engineering Research Center of AI and Robotics, Ministry of Education, China \and
Jilin Provincial Key Laboratory of Intelligence Science and Engineering, Changchun, China
}

\renewcommand{\thefootnote}{}
\footnotetext{$^{\ast}$Co-first author. $^{\textrm{\Letter}}$Corresponding author.}

\maketitle              
\begin{abstract}
While Large Language Models (LLMs) excel in world knowledge understanding, adapting them to specific subfields requires precise adjustments. Due to the model's vast scale, traditional global fine-tuning methods for large models can be computationally expensive and impact generalization. To address this challenge, a range of innovative Parameters-Efficient Fine-Tuning (PEFT) methods have emerged and achieved remarkable success in both LLMs and Large Vision-Language Models (LVLMs). In the medical domain, fine-tuning a medical Vision-Language Pretrained (VLP) model is essential for adapting it to specific tasks. Can the fine-tuning methods for large models be transferred to the medical field to enhance transfer learning efficiency?
In this paper, we delve into the fine-tuning methods of LLMs and conduct extensive experiments to investigate the impact of fine-tuning methods for large models on the existing multimodal model in the medical domain from the training data level and the model structure level. 
We show the different impacts of fine-tuning methods for large models on medical VLMs and develop the most efficient ways to fine-tune medical VLP models. We hope this research can guide medical domain researchers in optimizing VLMs' training costs, fostering the broader application of VLMs in healthcare fields.
The code and dataset have been released at \url{https://github.com/TIMMY-CHAN/MILE}.
\end{abstract}

\section{Introduction}
\label{sec:intro} 

The rise of ChatGPT has ignited significant interest in Large Language Models (LLMs). However, LLMs often require fine-tuning to adapt to specific domains such as medicine due to their general-purpose nature. Global fine-tuning methods are computationally expensive and may compromise model generalization capabilities. Therefore, numerous studies have begun exploring Parameter-Efficient Fine-Tuning (PEFT) techniques~\cite{p-tuning, GPT3, li2021prefix, lora} aimed at enabling more efficient fine-tuning and achieved remarkable success.

Some works~\cite{liu2023visual, chen2024efficiency} have attempted to apply PEFT methods to Large Vision-Language Models (LVLMs). Compared to Natural Language Processing (NLP) tasks, visual language tasks introduce visual inputs, leading to more diverse content and requiring more challenging fine-tuning. These endeavours have demonstrated that PEFT methods successful in LLMs can enhance the few-shot and zero-shot capabilities of LVLMs or achieve comparable results to global fine-tuning approaches.

However,  despite advancements~\cite{llama-adapter, li2023llava, minigpt} in reducing computational requirements for fine-tuning LVLMs, for many researchers, particularly those in interdisciplinary fields outside of Computer Science (CS), such as biomedicine, accessing the computational resources required for large-scale model fine-tuning remains a significant challenge. The lack of access to server-level GPUs, crucial for effective model fine-tuning, poses a considerable challenge. Consequently, there is a pressing need for small-scale VLMs (which are called basic or fundamental VLMs), particularly given the privacy concerns surrounding medical images. The legality and ethical concerns surrounding the upload of private medical images to publicly available LVLMs further emphasize the necessity of investigating whether PEFT methods, successful in LLMs and LVLMs, can achieve comparable results when applied to basic VLMs. 

To empower researchers in the medical domain with limited computational resources to effectively fine-tune multi-modal models for practical applications, we embark on experimental research to investigate the applicability of the LLMs' tuning methods in the realm of medical multimodal\,(vision-language) learning. In this paper, we design a \textbf{M}odularized med\textbf{I}cal vision-\textbf{L}anguage fine-tuning mod\textbf{E}l (\textbf{MILE}) that builds upon a medical Vision-Language Pretrained\,(VLP) model and incorporates various PEFT modules through modular design. Specifically, from the model structure perspective, we conduct a systematic investigation of the PEFT methods in the LLMs and develop the corresponding modules which can be integrated into a generative vision-language baseline model. From the training data perspective, we propose an instruction-format medical multi-modal dataset for applying instruction-tuning on different MILE variants. We conduct in-depth ablation studies on those LLMs' tuning methods and validate them on two radiographic image benchmarks. We believe that these empirical analyses will catalyze the development of fine-tuned medical multimodal models.

Our main contributions are as follows:
\textbf{(i)} We systematically explored how trainable parameters in different medical VLM modules affect overall performance, revealing strategies for achieving competitive results akin to global fine-tuning.
\textbf{(ii)} Through extensive experiments, we conducted a novel comparison of the PEFT methods tailored for small-scale medical VLM based on a baseline model, offering insights distinct from large-scale models.
\textbf{(iii)} We conducted a thorough analysis of the impact of instruction-tuning on fine-tuning basic VLP models and released an instruction-format medical image-text dataset. Our investigation revealed both positive and negative effects of instruction-tuning, offering a nuanced understanding of its implications for the fine-tuning process of the small-scale VLP models.

\section{Related Work}

\textbf{PEFT Techniques of LLMs:}
Fine-tuning large pretrained language models (PLMs) is resource-intensive, often requiring substantial computational resources and training data. To address this challenge, PEFT techniques have emerged, aiming to enhance PLMs' performance on specific tasks with minimal changes to model parameters. Various methods have been developed in this regard. Adapter Tuning~\cite{adapter_tuning}, Dora~\cite{liu2024dora}, LoRA~\cite{lora} etc.~\cite{IA3,p-tuningv2} add small components to PLMs or the input embedding~\cite{prompt, GPT3} to realize PEFT.
Besides methods that add small components, some PEFT techniques focus on data manipulation to minimize or eliminate changes to the original model weights. OpenAI~\cite{GPT3} and Google~\cite{instruction_tuning} have independently introduced instruction-tuning methods that modify original data into instruction pairs for fine-tuning models, achieving better generative results compared to multi-task training. 
While these PEFT methods have been successfully applied to LLMs, their impact on small-scale VLMs remains underexplored, especially in the medical domain.

\noindent \textbf{Medical Vision-Language Models:}
Recent advancements in the pretraining-finetuning paradigm have led to the emergence of medical VLMs~\cite{m3ae, miss, li2023masked} based on VLP models.  However, their pretraining and fine-tuning require data scales of more than 100,000 image-text pairs and the number of trainable parameters for global fine-tuning is not much different from that of PEFT in LVLMs. Therefore, under the dual factors of large-scale training data and high training parameters, small-scale VLMs' training costs remain unaffordable for many researchers. Thus, in this work,  we systematically review LLMs' tuning methods and discuss their applicability to medical VLMs.

\section{Method}
\subsection{Architecture of MILE}
\begin{figure}[]
    \centering
    \includegraphics[width=1\textwidth]{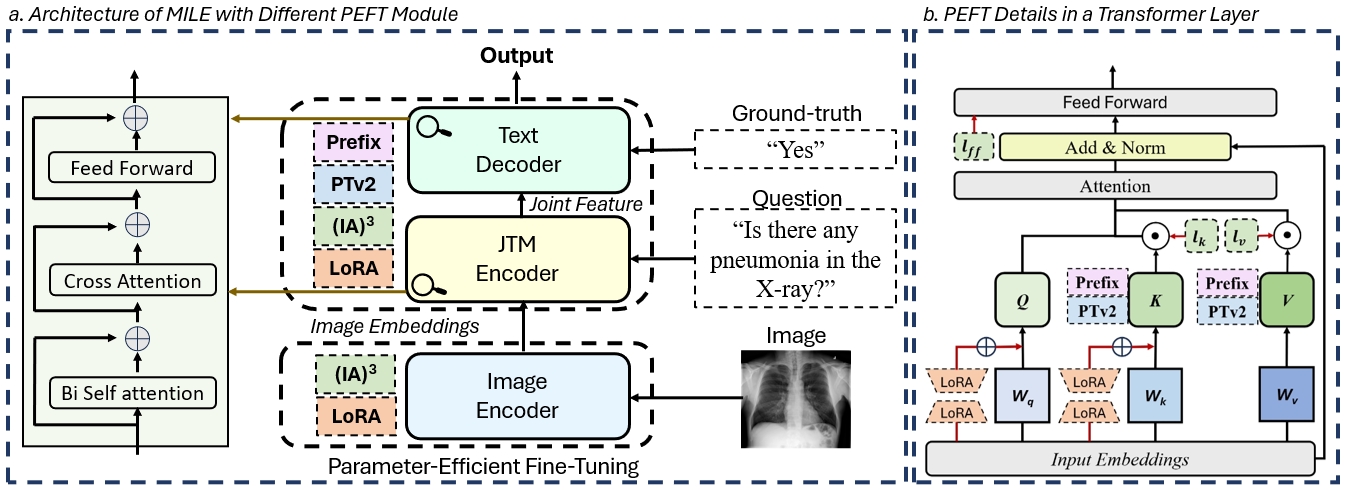}
    \caption{Architecture of MILE with different PEFT modules\,(a) and PEFT details in a Transformer layer\,(b).}
    \label{fig:example}
\label{fig1}
\end{figure}

\textbf{Baseline model:} Most VLMs architecture are based on CLIP~\cite{radford2021learning} or BLIP~\cite{li2022blip}. In this paper, we use MISS\cite{miss}, a generative multimodal medical VLM as our baseline model, the architecture has been shown in Figure \ref{fig1}a. MISS has an image encoder and a Joint Text-Multimodal\,(JTM) encoder, the former for image feature extraction and the latter for text feature extraction and multimodal feature interaction. A text decoder is appended after the JTM decoder for causal reasoning and text generation. The image encoder of the baseline model is a ViT-base\cite{vit} model; The JTM encoder is designed based on Bert with 12 transformer-based\cite{vaswani2017attention} layers where a cross-attention layer is inserted between the bi-self attention layer and the feed-forward layer; the architecture of the text decoder is similar to the JTM encoder and the bi-self attention layer is replaced by a causal attention layer.

\textbf{The Unified Model:} To validate the effectiveness of the above PEFT methods on small-scale medical VLMs, we construct MILE and equip it with 4 modules of commonly used PEFT methods~\cite{lora,li2021prefix, IA3, p-tuningv2}, obtaining four variants: MILE-LoRA, MILE-Prefix, MILE-IA3, and MILE-PTv2.

\setlength{\intextsep}{0pt}

\setlength{\intextsep}{0pt}

$\bullet$ MILE-LoRA: As shown in  Figure~\ref{fig1}b, low-rank matrixes are selectively injected into all the attention layers' parameter matrix $query$ and $key$ in the image encoder, JTM encoder or text decoder for LoRA-Tuning.

$\bullet$ MILE-$(IA)^3$: For $(IA)^3$-Tuning\,(IA3)~\cite{IA3}, learnable vectors $l_{k}$, $l_{v}$ and $l_{ff}$ which rescale the $key$, $value$ and the inner activations are respectively injected into all the attention layers and feed-forward layers in the image encoder, JTM encoder or text decoder of the baseline model during tuning, as shown in Figure~\ref{fig1}b. 

$\bullet$ MILE-Prefix: For MILE-Prefix, prefix vectors are selectively attached before the input of the JTM encoder or text decoder, while tuning the image encoder. When the prefix vectors are only attached before the input of the text decoder, the input embeddings are defined as $z = [PREFIX, x]$~\cite{li2021prefix}, while the prefix vectors are attached by both the input of the JTM encoder and text decoder the input is defined as $z = [PREFIX, x, PREFIX']$.

$\bullet$ MILE-PTv2: As shown in Figure~\ref{fig1}a, prompt tokens are selectively attached before the input of the JTM encoder or text decoder. Different from prefix-tuning, all prefixes in the input of each attention layer are derived from a trainable matrix when we apply P-Tuning v2\cite{p-tuningv2}. 
\subsection{Instruction-format Data Generation}

To analyze the impact of instruction-tuning on basic VLMs, we curated a medical image-text dataset using Instruction templates (details in the Appendix). 'Closed' templates suit closed-ended questions, and 'Opened' templates are for open-ended ones. During training, each QA pair randomly incorporates a template. For answer options, inspired by~\cite{miss}, we categorized question attributes and created diverse candidate pools. In opened instructions, incorrect answers from the same attribute are randomly included with the correct answer.

\subsection{Training}

We use the Slake Dataset\cite{liu2021slake} and the VQA-RAD Dataset~\cite{lau2018dataset} for training, testing and validating. To ensure a fair comparison of the impact of instruction-tuning and other PEFT methods on basic VLMs, the setting of the training hyper-parameter is the same with \cite{miss} and the dataset splits are identical to those used in most current works~\cite{m3ae, miss,li2023masked}. The training loss $\mathcal{L}$ is the language modeling loss~\cite{devlin2018bert}. For each MILE employing PEFT, the PEFT parameters are varied during training (if applicable) to investigate the effect of different percentage parameter changes on model performance. Each MILE is trained with both instruction-format data we make and original data provided by the dataset, getting two different models. Testing is conducted under the original benchmark and the task is generative, with no candidate answers provided to the model. 
\section{Experiment Results and Analysis}
\label{sec4}
We initially trained a series of MILE equipped with PEFT modules using the original data. Tables~\ref{tab1} to \ref{tab4} present the accuracy\,(ACC(\%)) of four MILE variants training with the origin data on the Slake benchmark. 'F' denotes the freezing of all parameters within a given module, 'T' signifies that all parameters are trainable, and the acronyms 'LoRA', 'Prefix', 'IA3', and 'PTV2' identify the specific PEFT methods applied. 'Memory' represents the GPU memory\,(GB) required for training. And '\#Params' indicates the weight of trainable parameters over all parameters.
\begin{table}[]
\setlength{\tabcolsep}{1pt}
\centering
\caption{Results of MILE-LoRA(origin data).}
\begin{tabular}{lll|ccc|ccc}
\hline
ViT                 & JTM                    & Dec  & Rank & \#Params   & Memory                     & Opened & Closed & Gobal \\ \hline
                    &                        & LoRA & 4    &0.163\%     & 5.19                            & 3.57   & 50.70   & 20.34 \\
\multirow{-2}{*}{F} & \multirow{-2}{*}{LoRA} & LoRA & 8    &0.325\%     & 5.21                            & 3.57   & 50.70   & 20.34 \\ \hline
                    &                        &      & 4    & 0.327\%    & 26.63                           & 48.65  & 50.70   & 49.34 \\
\multirow{-2}{*}{LoRA} & \multirow{-2}{*}{LoRA} & \multirow{-2}{*}{LoRA} & 8 & 0.652\%   & 26.75                              & 48.93 & 50.70  & 49.57 \\ \hline
                    &                        & LoRA & 4    &  38.022\%    & 7.26                           & 47.76  & 70.70   & 55.53 \\
\multirow{-2}{*}{F}    & \multirow{-2}{*}{T}    & LoRA     & 8 &  38.072\%    &  7.45                         & 50.21 & 70.99 & 57.18 \\ \hline
                    &                        & LoRA & 4    &   24.009\%     &  26.96                        & 68.14  & 50.70   & 62.29 \\
\multirow{-2}{*}{T} & \multirow{-2}{*}{LoRA} & LoRA & 8    &   24.133\%    &   27.29                         & 68.28  & 50.70   & 62.38 \\ \hline
                    &                        &      & 4    & 61.887\%      &  27.60               & 78.52  & 79.44  & 78.83 \\
\multirow{-2}{*}{T}    & \multirow{-2}{*}{T}    & \multirow{-2}{*}{LoRA} & 8 & 61.919\%      &  28.11                & 78.66 & 80.56 & 79.30 \\ \hline
\end{tabular}
\label{tab1}
\end{table}
\setlength{\intextsep}{1pt}

The results demonstrate that a fully frozen visual encoder\,(ViT) within the VLM significantly hampers the model's ability to correctly interpret texts and images, as observed in the MILE-IA3 and MILE-PTV2, where global ACC plummets to 0.57\% and 0\%, respectively. Conversely, a modest increase in tunable parameters by 0.16\% in MILE-LoRA leads to notable improvements of 45\% and 29\% in open-ended and global ACC, respectively.  When all parameters of the ViT are set to trainable, the global ACC of the four models increases by 42\%, 21\%, 17\%, and 19\% compared to when they are completely frozen.

\begin{figure}
  \begin{minipage}{0.55\textwidth}
    \begin{minipage}{\linewidth}
      \centering
      \captionof{table}{Results of MILE-Prefix.}
      \begin{adjustbox}{max width=\linewidth}
        \begin{tabular}{ccc|cc|ccc}
          \hline
          ViT & JTM & Dec & \#Params & Memory & Opened & Closed & Global \\ \hline
          F   & F   & Prefix & 3.926\% & 4.62 & 0     & 50.7  & 17.3  \\
          F   & Prefix & Prefix & 7.556\% & 4.67 & 0     & 50.7  & 17.3  \\
          T   & Prefix & Prefix & 29.636\% & 26.41 & 41.50  & 32.95 & 38.61 \\
          T   & T   & Prefix & 63.354\% & 27.97 & 76.82 & \textbf{82.25} & 78.65 \\ \hline
        \end{tabular}
        \label{tab2}
      \end{adjustbox}
    \end{minipage}
    \setlength{\intextsep}{2pt}
    \begin{minipage}{\linewidth}
      \centering
      \setlength{\intextsep}{2pt}
      \captionof{table}{Results of MILE-IA3.}
      \begin{adjustbox}{max width=\linewidth}
        \begin{tabular}{lll|cc|ccc}
          \hline
          ViT & JTM & Dec & \#Params & Memory & Opened & Closed & Global \\ \hline
          F   & IA3 & IA3 & 0.051\%  & 6.35     & 0      & 1.69   & 0.57  \\
          IA3 & IA3 & IA3 & 0.061\%  & 23.01    & 0      & 50.70  & 16.98 \\
          T   & IA3 & IA3 & 23.924\% & 26.83     & 12.77  & 28.17  & 17.92 \\
          F   & T   & IA3 & 37.987\% & 7.52     & 46.24  & 50.70  & 47.74 \\
          T   & T   & IA3 & 61.866\% & 27.90     & 72.20  & 47.04  & 63.77 \\ \hline
        \end{tabular}
        \label{tab3}
      \end{adjustbox}
    \end{minipage}
    \begin{minipage}{\linewidth}
      \centering
      \captionof{table}{Results of MILE-PTV2.}
      \begin{adjustbox}{max width=\linewidth}
        \begin{tabular}{lll|cc|ccc}
          \hline
          ViT & JTM  & Dec  & \#Params & Memory & Opened & Closed & Global \\ \hline
          F   & PTV2 & PTV2 & 0.102\%  & 4.52    & 0      & 0      & 0     \\
          F   & F    & PTV2 & 0.051\%  & 4.57    & 7.10   & 0      & 4.72  \\
          T   & PTV2 & PTV2 & 23.963\% & 25.41     & 13.62  & 29.30  & 18.87 \\
          T   & T    & PTV2 & 61.876\% & 27.46     & 74.18  & 49.86  & 66.04 \\ \hline
        \end{tabular}
        \label{tab4}
      \end{adjustbox}
    \end{minipage}
  \end{minipage}%
  \begin{minipage}{0.45\textwidth}
    \begin{minipage}{\linewidth}
      \centering
      \includegraphics[width=0.71\linewidth]{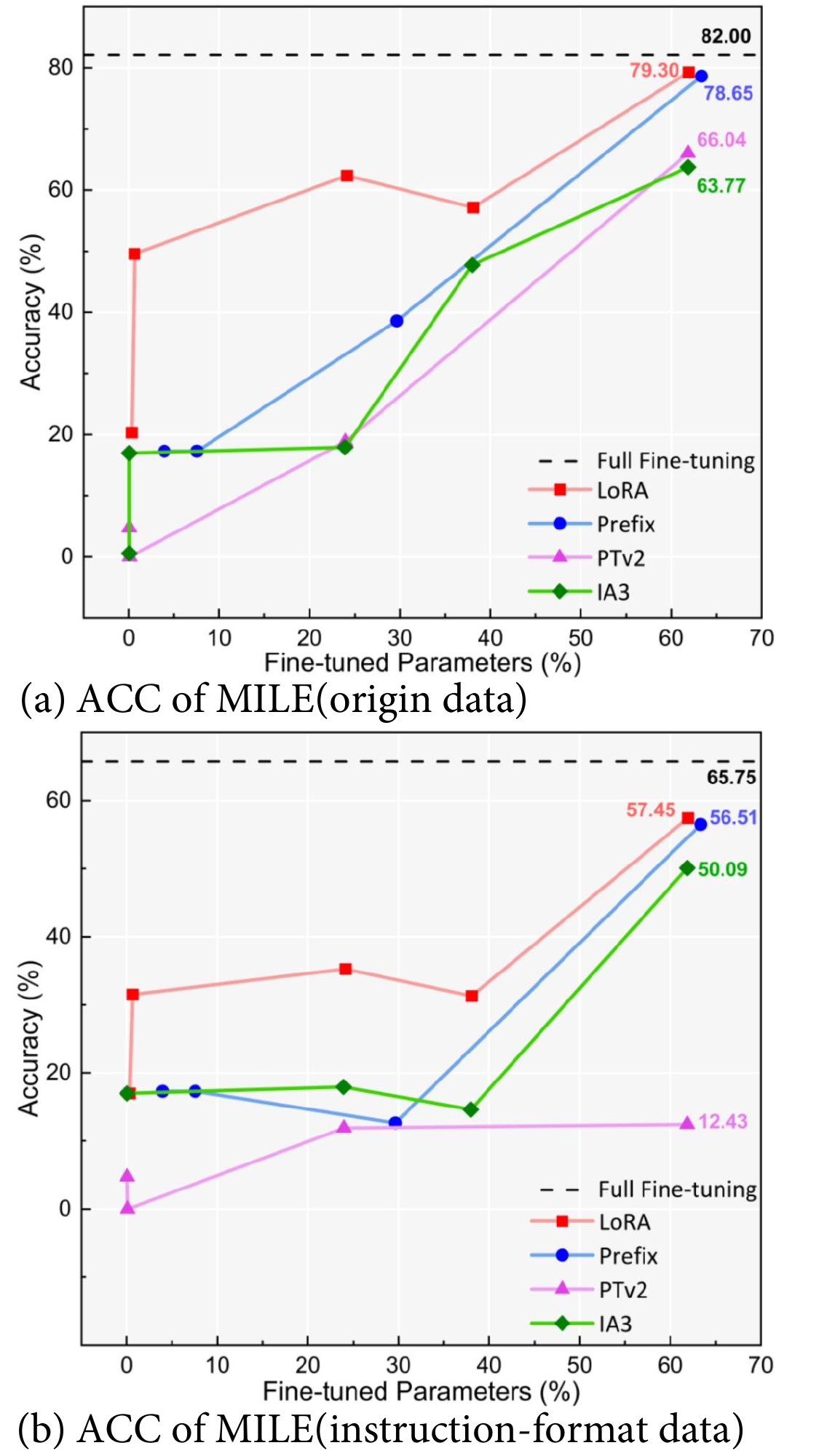}
      \captionof{figure}{ACC of MILE trained with different data.}
      \label{fig2}
    \end{minipage}%
  \end{minipage}
\end{figure}

On the other hand, in MILE-Prefix and MILE-IA3, when the ViT is frozen, even converting the JTM encoder from frozen to fully trainable barely improves global ACC. This underscores the pivotal role of the visual encoder in a VLP model for downstream task adaptation, where even minimal adjustments via PEFT can significantly enhance performance.

When the parameters of the ViT are updatable, increasing the proportion of parameter updates for the JTM encoder can also lead to significant improvements. Notably, in MILE-Prefix, shifting from Prefix-Tuning to full parameter updates for the JTM encoder boosts global ACC by 40\%, with closed-ended question ACC surpassing that of the baseline model employing global fine-tuning. In MILE-IA3 and MILE-PTV2, elevating the update ratio for the JTM encoder markedly improves open-ended question ACC by 67\% and 70\%. However, this comes with the cost of increasing the training parameter ratio to 61\% to 64\%.

It is also worth mentioning that full parameters updating of both the visual encoder and JTM encoder, alongside PEFT application to the decoder, can reduce the parameter count by 40\% while maintaining performance on par with global fine-tuning.

\subsection{Performance Differences Among Different PEFT Methods}
Although the aforementioned PEFT methods have been compared in their respective papers within the LM domain, our experiments reveal differing efficacies of them within the medical VLM domain. 

LoRA-Tuning exhibits the most competitive performance in this domain, effective for PEFT both language modeling Transformers and visual modeling Transformers. Compared with IA3, when all model parameters are subject to PEFT, MILE-IA3 consistently answers 'no'\,(ACC 50.70\%) to all questions. While MILE-LoRA also responds 'no' to closed-ended questions, it demonstrates a better understanding of the semantic information from images and text, achieving a 48.65\% ACC on open-ended questions, outperforming MILE-Prefix, MILE-IA3, and MILE-PTV2, which have about 20\%-30\% trainable parameters.

Furthermore, we conducted an ablation study on the rank of the LoRA unit. As shown in Table \ref{tab1}, within the same tuning paradigm, a doubling of the number of parameters in the LoRA matrix brings about a minor improvement in model performance. A LoRA rank of 8\,(ViT froze) and LoRA-tuned\,(rank=4) visual encoders, adjusting a similar parameter fraction (about 0.32\%), differed by 45\% in open-ended questions' accuracy.

MILE-Prefix also demonstrates promising results. When both the ViT and the JTM are fully trainable, and the decoder employs PEFT, the performance of MILE-Prefix is comparable to MILE-LoRA. When the JTM encoder also utilizes PEFT, MILE-Prefix lags, indicating that simply adding a prefix to vectors does not effectively promote the alignment of features from different modalities in the cross-attention layer, consistent with the principle of Prefix-Tuning.

Compared to the baseline model, the performance of MILE-IA3 and MILE-PTV2 is inferior to MILE-LoRA and MILE-Prefix. Within fundamental VLM, updating a negligible fraction\,(less than 0.1\%) of the decoder's parameters substantially impairs its generative task performance. This impact is markedly less pronounced in LLMs, underscoring the critical role of the number of parameters proportionality in tuning efficacy.

\begin{figure}
    \centering
    \includegraphics[width=1\linewidth]{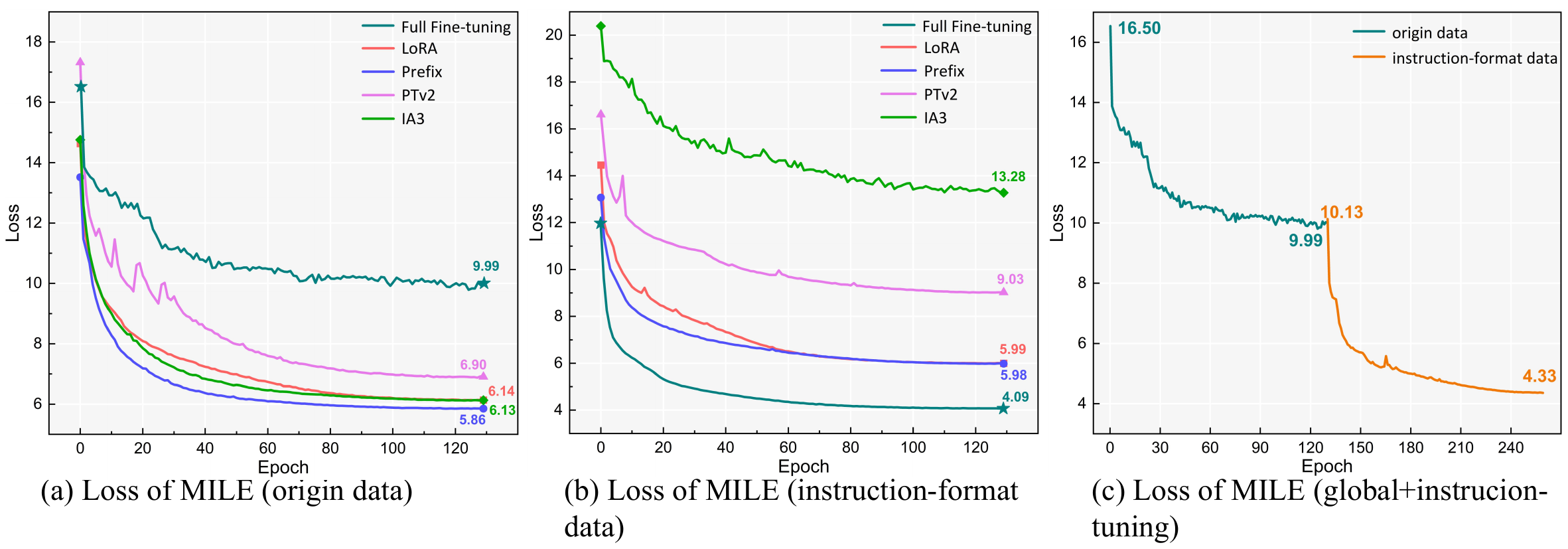}
    \caption{Loss of MILE variants under the training with origin data\,(a), instruction-format data\,(b) and an instruction-tuning MILE after fine-tuning by origin data\,(c).}
    \label{fig3}
\end{figure}

\subsection{The Impact of Data on Training Effectiveness} 
We employed instruction-format image-text pairs to conduct both global fine-tuning and PEFT of the model but did not parallel the positive effects seen in LLMs when applied to a basic VLP model. As shown in Table \ref{tab5}, the use of instruction-format data during full fine-tuning led to more than 20\% decrease in the global ACC. Moreover, Figure \ref{fig2}a\&b shows that the synergy of structure-level PEFT with data-level instruction-tuning significantly reduced the performance of various MILE variants by over 20\%, diverging from outcomes with raw data\,(more precise data are presented in tables in the Appendix).

This disparity underscores the potential model and task-specific sensitivity of instruction-format data's benefits. For basic VLP models, such data application might not yield the expected advantages.

\subsection{Is Instruction-Tuning Ineffective for Basic VLMs?}
While LLMs and LVLMs can benefit from instruction-tuning to enhance their generalization across different types of tasks, this approach may not be effective for basic VLMs. This is because base VLMs are typically fine-tuned for specific downstream tasks. Furthermore, instruction-format text inputs provide candidate answers to questions\,(see in Appendix), which do not exist in real-world scenarios during inference and practical applications of generative models.

As shown in Figure \ref{fig3}a\&b, global fine-tuning with instruction-based data showed lower losses, hinting at an improved task format comprehension but potentially oversimplifying the difficulty of training tasks, thus diminishing real-world inferencing efficacy. 

However, given that instruction-format data can enhance the model's understanding of the target task and result in lower training losses, could it potentially improve a global fine-tuned model that has already converged on the original data? Figure \ref{fig3}c demonstrates the loss reduction when a converged model is further trained on instruction-format data, the loss of the converged model continues to decrease and finally reaches about 4.33 after global fine-tuning. As shown in Table \ref{tab5}, MILE ultimately achieves 86.7\% open-ended ACC, 81.42\% closed-ended ACC, and 83.02\% overall ACC, surpassing the baseline model and demonstrating state-of-the-art performance among generative VLMs.

\begin{table*}[htbp]
\caption{Comparsion with other medical VLMs which have different tuning paradigms, '$\clubsuit$' means global fine-tuning with ordinary data\,(no instruction-format) and '$\spadesuit$' means instruction-tuning with all parameters updating.}
\resizebox{\linewidth}{!}{
\begin{tabular}{lccccccccc}
\hline
\multirow{2}{*}{Methods} &
  \multicolumn{1}{c|}{\multirow{2}{*}{\begin{tabular}[c]{@{}c@{}}Pretrain\\ \# images\end{tabular}}} &
  \multirow{2}{*}{\begin{tabular}[c]{@{}c@{}}Tuning\\ paradigm\end{tabular}} &
  \multicolumn{1}{c|}{\multirow{2}{*}{Type of task}} &
  \multicolumn{3}{c|}{VQA-RAD} &
  \multicolumn{3}{c}{SLALKE} \\ 
 &
  \multicolumn{1}{c|}{} &
   &
  \multicolumn{1}{c|}{} &
  CLOSED &
  OPENED &
  \multicolumn{1}{c|}{OVERALL} &
  CLOSED &
  OPENED &
  OVERALL \\ \hline

MTL\cite{cong2022caption} &
  \multicolumn{1}{c|}{87,952} &
  $\clubsuit$ &
  \multicolumn{1}{c|}{classification} &
  79.8 &
  69.8 &
  \multicolumn{1}{c|}{75.8} &
  86.1 &
  80.2 &
  82.5 \\
M3AE\cite{m3ae} &
  \multicolumn{1}{c|}{298,000} &
  $\clubsuit$ &
  \multicolumn{1}{c|}{classification} &
  83.4 &
  67.2 &
  \multicolumn{1}{c|}{77} &
  87.8 &
  80.3 &
  83.2 \\
MUMC\cite{li2023masked} &
  \multicolumn{1}{c|}{387,000} &
  $\clubsuit$ &
  \multicolumn{1}{c|}{ranking} &
  84.2 &
  71.5 &
  \multicolumn{1}{c|}{79.2} &
  - &
  - &
  84.9 \\
MISS\cite{miss} &
  \multicolumn{1}{c|}{38,800} &
  $\clubsuit$ &
  \multicolumn{1}{c|}{\textbf{generating}} &
  80.35 &
  \textbf{71.81} &
  \multicolumn{1}{c|}{76.05} &
  82.91 &
  \textbf{81.47} &
  82 \\ 
  \textbf{MILE} &
  \multicolumn{1}{c|}{38,800} &
  $\spadesuit$ &
  \multicolumn{1}{c|}{\textbf{generating}} &
  2.68 &
  45.58 &
  \multicolumn{1}{c|}{24.22} &
   59.72&
  68.79 &
  65.75 \\ 
  \textbf{MILE} &
  \multicolumn{1}{c|}{38,800} &
  $\clubsuit+\spadesuit$ &
  \multicolumn{1}{c|}{\textbf{generating}} &
  76.34 &
  \textbf{73.45} &
  \multicolumn{1}{c|}{74.89} &
  \textbf{86.70} &
  81.42 &
  \textbf{83.02} \\ \hline
\end{tabular}
}
\label{tab5}
\end{table*}
\section{Conclusion}

In this paper, we comprehensively investigate whether fine-tuning methods for LLMs can be applied to the medical multimodal domain, aiming to ease the training burden on resource-constrained practitioners. We developed a suite of MILE models incorporating various fine-tuning strategies atop generative VLP frameworks, delving into the effects of structural and parametric modifications on performance.
From a series of experiments, we observe that updating the parameters of the visual encoder is crucial for VLMs. Furthermore, updating the parameters of the JTM encoder which is responsible for text feature extraction and multimodal feature fusion can significantly enhance model performance. Through a comparison of different PEFT methods, we find that LoRA-Tuning and Prefix-Tuning exhibit the best tuning effects, achieving comparable performance to global fine-tuning models while reducing training costs by 40\%.

Additionally, we explore the impact of data-level fine-tuning, specifically instruction-tuning, on model performance. Although directly fine-tuning with instruction-format data simplifies the training task, it leads to suboptimal performance for basic VLMs in practical tasks. Nonetheless, instruction-tuning on top of models already optimized on original datasets demonstrated notable performance gains.
We hope that our work can inspire researchers in the medical field who aim to reduce the training costs of multimodal models and promote the application of VLMs in the medical domain.

\section{Acknowledgement}
This work is supported in part by the National Key R\&D Program of China (2021ZD0113502).

\newpage
\bibliographystyle{splncs04}

\begin{thebibliography}{10}
\providecommand{\url}[1]{\texttt{#1}}
\providecommand{\urlprefix}{URL }
\providecommand{\doi}[1]{https://doi.org/#1}

\bibitem{GPT3}
Brown, T., Mann, B., Ryder, et~al.: Language models are few-shot learners. Advances in neural information processing systems  \textbf{33},  1877--1901 (2020)

\bibitem{chen2024efficiency}
Chen, J., Yang, D., Jiang, Y., Li, M., Wei, J., Hou, X., Zhang, L.: Efficiency in focus: Layernorm as a catalyst for fine-tuning medical visual language pre-trained models. arXiv preprint arXiv:2404.16385  (2024)

\bibitem{miss}
Chen, J., Yang, D., Jiang, Y., et~al.: Miss: A generative pretraining and finetuning approach for med-vqa. arXiv preprint arXiv:2401.05163  (2024)

\bibitem{m3ae}
Chen, Z., Du, Y., Hu, et~al.: Multi-modal masked autoencoders for medical vision-and-language pre-training. In: MICCAI. pp. 679--689. Springer (2022)

\bibitem{cong2022caption}
Cong, F., Xu, S., et~al.: Caption-aware medical vqa via semantic focusing and progressive cross-modality comprehension. In: ACM MM. pp. 3569--3577 (2022)

\bibitem{devlin2018bert}
Devlin, J., Chang, M.W., Lee, et~al.: Bert: Pre-training of deep bidirectional transformers for language understanding. arXiv preprint arXiv:1810.04805  (2018)

\bibitem{vit}
Dosovitskiy, A., Beyer, L., Kolesnikov, A., Weissenborn, D., Zhai, X., Unterthiner, T., Dehghani, M., Minderer, M., Heigold, G., Gelly, S., et~al.: An image is worth 16x16 words: Transformers for image recognition at scale. arXiv preprint arXiv:2010.11929  (2020)

\bibitem{adapter_tuning}
Houlsby, N., Giurgiu, A., Jastrzebski, et~al.: Parameter-efficient transfer learning for nlp. In: International Conference on Machine Learning. pp. 2790--2799. PMLR (2019)

\bibitem{lora}
Hu, E.J., Shen, et~al.: Lora: Low-rank adaptation of large language models. arXiv preprint arXiv:2106.09685  (2021)

\bibitem{lau2018dataset}
Lau, J.J., Gayen, et~al.: A dataset of clinically generated visual questions and answers about radiology images. Scientific data  \textbf{5}(1),  1--10 (2018)

\bibitem{prompt}
Lester, B., Al-Rfou, R., Constant, N.: The power of scale for parameter-efficient prompt tuning. arXiv preprint arXiv:2104.08691  (2021)

\bibitem{li2023llava}
Li, C., Wong, C., Zhang, et~al.: Llava-med: Training a large language-and-vision assistant for biomedicine in one day. arXiv preprint arXiv:2306.00890  (2023)

\bibitem{li2022blip}
Li, J., Li, D., Xiong, et~al.: Blip: Bootstrapping language-image pre-training for unified vision-language understanding and generation. In: ICCV. pp. 12888--12900 (2022)

\bibitem{li2023masked}
Li, P., Liu, G., He, et~al.: Masked vision and language pre-training with unimodal and multimodal contrastive losses for medical visual question answering. In: MICCAI. pp. 374--383. Springer (2023)

\bibitem{li2021prefix}
Li, X.L., Liang, P.: Prefix-tuning: Optimizing continuous prompts for generation. arXiv preprint arXiv:2101.00190  (2021)

\bibitem{liu2021slake}
Liu, B., Zhan, L.M., Xu, et~al.: Slake: A semantically-labeled knowledge-enhanced dataset for medical visual question answering. In: 2021 ISBI. pp. 1650--1654 (2021)

\bibitem{IA3}
Liu, H., Tam, D., Muqeeth, et~al.: Few-shot parameter-efficient fine-tuning is better and cheaper than in-context learning. Advances in Neural Information Processing Systems  \textbf{35},  1950--1965 (2022)

\bibitem{liu2023visual}
Liu, H., Li, C., Wu, et~al.: Visual instruction tuning. arXiv preprint arXiv:2304.08485  (2023)

\bibitem{liu2024dora}
Liu, S.Y., Wang, C.Y., Yin, H., Molchanov, P., Wang, Y.C.F., Cheng, K.T., Chen, M.H.: Dora: Weight-decomposed low-rank adaptation. arXiv preprint arXiv:2402.09353  (2024)

\bibitem{p-tuningv2}
Liu, X., Ji, K., Fu, Y., Tam, et~al.: P-tuning v2: Prompt tuning can be comparable to fine-tuning universally across scales and tasks. arXiv preprint arXiv:2110.07602  (2021)

\bibitem{p-tuning}
Liu, X., Zheng, Y., Du, Z., Ding, et~al.: Gpt understands, too. AI Open  (2023)

\bibitem{radford2021learning}
Radford, A., Kim, J.W., Hallacy, et~al.: Learning transferable visual models from natural language supervision. In: International conference on machine learning. pp. 8748--8763. PMLR (2021)

\bibitem{vaswani2017attention}
Vaswani, A., Shazeer, N., Parmar, et~al.: Attention is all you need. NIPS  \textbf{30} (2017)

\bibitem{instruction_tuning}
Wei, J., Bosma, M., Zhao, V.Y., Guu, et~al.: Finetuned language models are zero-shot learners. arXiv preprint arXiv:2109.01652  (2021)

\bibitem{llama-adapter}
Zhang, R., Han, J., Zhou, A., Hu, X., Yan, S., Lu, P., Li, H., Gao, P., Qiao, Y.: Llama-adapter: Efficient fine-tuning of language models with zero-init attention. arXiv preprint arXiv:2303.16199  (2023)

\bibitem{minigpt}
Zhu, D., Chen, J., Shen, X., Li, X., Elhoseiny, M.: Minigpt-4: Enhancing vision-language understanding with advanced large language models. arXiv preprint arXiv:2304.10592  (2023)

\end{thebibliography}

\newpage
\section*{Appendix}
\renewcommand{\thesubsection}{\arabic{subsection}}
\vspace{-10pt}
\subsection{Templates of instruction-format data}
\begin{figure}[]
    \centering
    \includegraphics[width=0.95\textwidth]{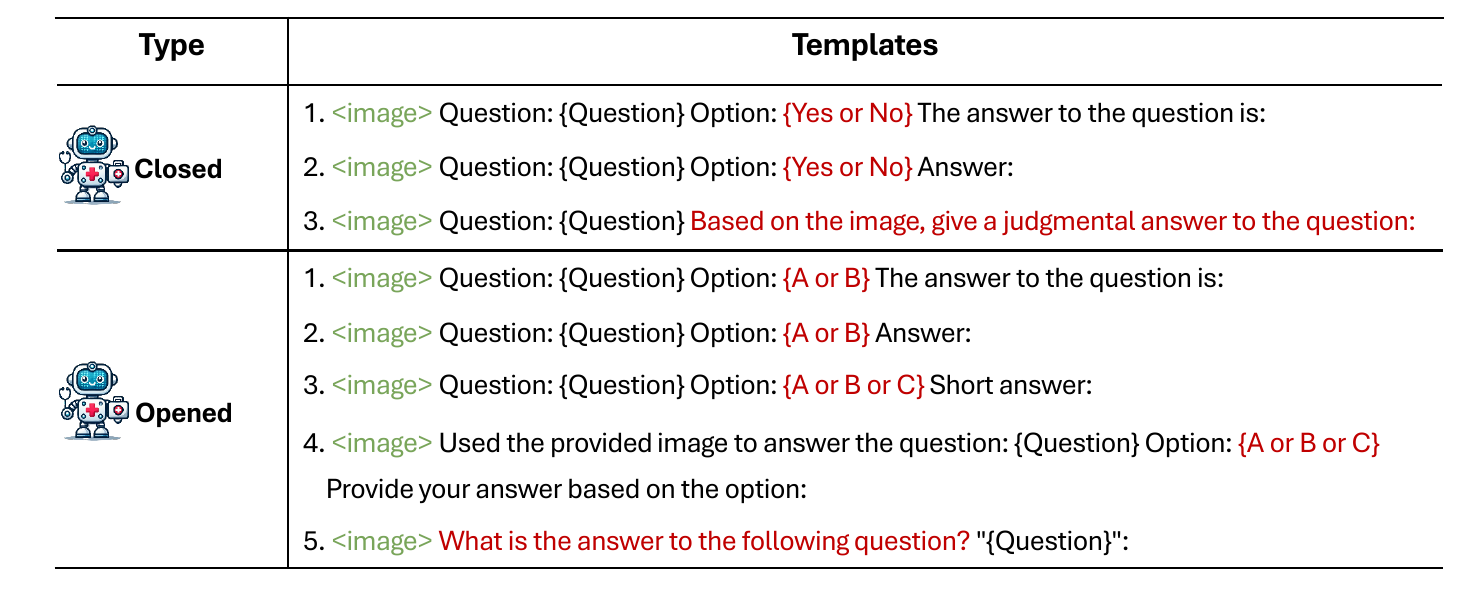} 
    \vspace{-\baselineskip}

    \caption{Instruction-format Data Templates.}
    \label{fig:1}
\end{figure}

\begin{figure}[]
    \centering
    \includegraphics[width=0.95\textwidth]{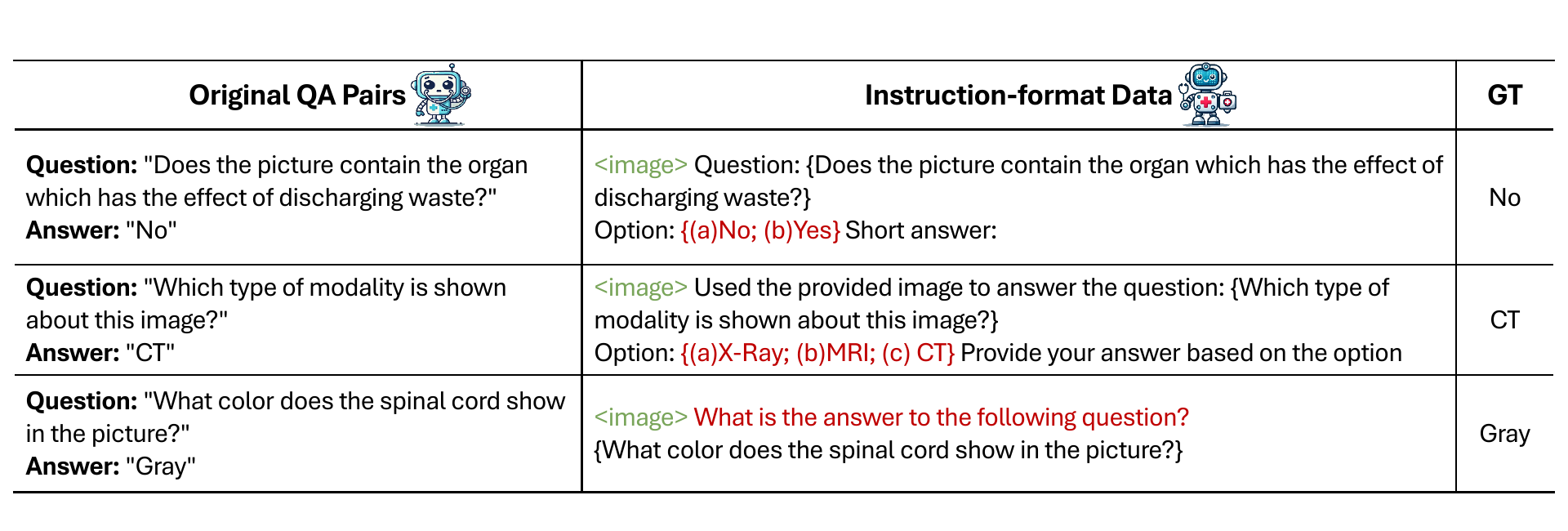} 
    \vspace{-\baselineskip}

    \caption{Instruction-format Data Examples.}
    \label{fig:2}
    \vspace{-5pt}
\end{figure}

\vspace{-5pt}
We constructed an instruction-format medical image-text dataset to investigate their impact on fundamental VLMs. In detail, we transformed the QA pairs into an instruction fine-tuning format by constructing instruction templates, as shown in Figures~\ref{fig:1} to \ref{fig:2}. 'Closed' and 'Opened' templates are designed for closed-ended and open-ended questions respectively. Each QA pair is randomly embedded with one of these templates during training. To generate answer options, we classified the question attributes based on modality, plane, shape, size, organ, location, and pathology. Consequently, we created pools of candidate answers for different question attributes. Incorrect answers from the same attribute are randomly selected and embedded into the option together with the ground-truth answer.
\setlength{\intextsep}{10pt}
\vspace{-20pt}

\vspace{-\baselineskip}

\subsection{Details of Experiments}
\vspace{-3pt}
\textbf{More Detailed Results of MILE Variants}:
As mentioned in Section 4 of our paper, we trained a series of MILE variants using the instruction-format data, and Tables \ref{tab:APX-1} to \ref{tab:APX-2} present the related results of four MILE variants on the Slake dataset. 
To more fully demonstrate the impact of different PEFT methods on the basic visual language model, we have also verified the PEFT method's impacts on BiomedGPT-Tiny: only the decoder is fine-tuned with PEFT methods, and the rest is full parameters fine-tuned. Table~\ref{tab:APX-3} presents the ACC on the benchmark, which shows a similar trend as MILE. 

\textbf{Implement Details}: Here, we will present the experimental details of our model. All the training was conducted on a single NVIDIA RTX8000-48GB GPU. We used the Adamw optimizer with cosine learning rate decay, an initial learning rate of 2e-5, a weight decay of 0.05, and a minimum learning rate of 0. The input image size for our model was 480$\times$480 pixels, trained for 130 epochs, with a batch size of 20. For our baseline model, the Vit-based visual encoder consists of 12 layers of transformer, and both the JTM encoder and text decoder contain 12 layers of transformer-based layers.

\textbf{Why choose this baseline model:} We chose MISS as our baseline model and developed MILE based on it because MISS is a small-scale generative VLM and it has similar architecture to current LVLMs such as LLaVA, BLIP2, and et.al. MISS unifies the Text encoder and the Multimodal encoder by a JTM encoder so we can more easily fine-tune it. 

\vspace{-\baselineskip}
\begin{table}[]
\centering
\footnotesize
\caption{Results of MILE-LoRA\,(instruction-format data).}
\resizebox{0.65\textwidth}{!}{
\begin{tabular}{lll|ccc|ccc}
\hline
ViT                 & JTM                    & Dec  & Rank & \#Params & Memory                           & Opened & Closed & Gobal \\ \hline
                    &                        & LoRA & 4    &0.163\%   & 5.44                              & 0   & 50.7   & 16.98 \\
\multirow{-2}{*}{F} & \multirow{-2}{*}{LoRA} & LoRA & 8    &0.325\%   & 5.53                              & 0   & 50.7   & 16.98 \\ \hline
                    &                        &      & 4    & 0.327\%  & 26.90                              & 28.09  & 29.01   & 28.40 \\
\multirow{-2}{*}{LoRA} & \multirow{-2}{*}{LoRA} & \multirow{-2}{*}{LoRA} & 8 & 0.652\%  & 27.01                              & 48.93 & 24.82  & 31.51 \\ \hline
                    &                        & LoRA & 4    &  38.022\%        & 7.85                       & 21.79  & 35.77   & 26.49 \\
\multirow{-2}{*}{F}    & \multirow{-2}{*}{T}    & LoRA     & 8 &  38.072\%    & 7.94                          & 27.35 & 39.44 & 31.32 \\ \hline
                    &                        & LoRA & 4    &   24.009\%       & 26.95                       & 39.57  & 8.73   & 29.25 \\
\multirow{-2}{*}{T} & \multirow{-2}{*}{LoRA} & LoRA & 8    &   24.133\%       & 27.62                        & 41.42  & 23.10   & 35.28 \\ \hline
                    &                        &      & 4    & 61.887\%         & 27.66             & 61.28  & 36.34  & 52.92 \\
\multirow{-2}{*}{T}    & \multirow{-2}{*}{T}    & \multirow{-2}{*}{LoRA} & 8 & 61.919\%     & 28.63                  & 63.54 & 45.35 & 57.45 \\ \hline
\end{tabular}}
\label{tab:APX-1}
\vspace{-20pt}
\end{table}

\vspace{2pt}
\begin{table}[]
\centering
\footnotesize
\caption{Results of MILE-Prefix \& IA3 \& PTV2\,(instruction-format data).}
\resizebox{0.65\textwidth}{!}{
\begin{tabular}{lll|cc|ccc}
\hline
ViT                 & JTM                    & Dec  & \#Params   & Memory                        & Opened & Closed & Gobal \\ \hline
F   & IA3 & IA3 &  0.051\%  & 6.41     & 0      & 50.70   & 16.98  \\
IA3 & IA3 & IA3 &  0.061\%  & 23.47   & 0      & 50.70  & 16.98 \\
T   & IA3 & IA3 &  23.924\% & 27.42     & 12.77  & 27.04  & 17.92 \\
F   & T   & IA3 &  37.987\% & 8.18     & 8.37  & 27.04  & 14.62 \\
T   & T   & IA3 &  61.866\% & 28.81     & 50.21  & 49.86  & 50.09 \\ \hline

F & F      & Prefix & 3.926\% & 4.76 & 0     & 50.70  & 17.30  \\
F & Prefix & Prefix & 7.556\% & 4.81 & 0     & 50.70  & 17.30  \\
T & Prefix & Prefix & 29.636\%& 26.56  & 7.23  & 22.38 & 12.64 \\
T & T      & Prefix & 63.354\%& 28.14  & 68.65 & 32.39 & 56.51 \\
 \hline

F   & F    & PTV2 &  0.051\%  & 4.66     & 7.10   & 0      & 4.72 \\
F   & PTV2 & PTV2 &  0.102\%  & 4.70     & 0      & 0      & 0   \\
T   & PTV2 & PTV2 &  23.963\% & 26.03     & 6.10  & 23.38  & 11.89 \\
T   & T    & PTV2 &  61.876\% & 27.83     & 3.12  & 30.99  & 12.43 \\ \hline
\end{tabular}}
\label{tab:APX-2}
\end{table}
\vspace{-\baselineskip}

\setlength{\intextsep}{15pt}

\begin{table}[]
\centering
\footnotesize

\caption{Results on BiomedGPT-Tiny\,(origin data).}
\resizebox{0.6\textwidth}{!}{
\begin{tabular}{l|cccc}
\hline
Method                      & \#Params  & Opened & Closed & Global \\ \hline
Full Fine-tuning            & 100\%     & 71.84  & 64.46  & 68.97  \\
Decoder-LoRA                & 50.76\%   & 66.82  & 63.48  & 65.52  \\
Decoder-Prefix              & 51.05\%   & 69.94  & 60.54  & 66.29  \\
Decoder-IA3                 & 50.49\%   & 64.95  & 52.21  & 60.01  \\
Decoder-PTV2                & 50.92\%   & 68.07  & 48.78  & 60.57  \\ \hline
\end{tabular}}
\label{tab:APX-3}
\end{table}
\end{document}